\documentclass[conference]{IEEEtran}

\usepackage{cite}
\usepackage{graphicx}
\usepackage{array}
\usepackage{amsmath}
\usepackage{algorithm}
\usepackage{algorithmic}
\usepackage{subfigure}
\usepackage{soul}
\usepackage{url}
\usepackage{multirow,makecell}
\usepackage{diagbox,booktabs,mdwlist}
\usepackage{bm}

\newcommand{\etal}{\textsl{et~al.}}

\newcommand{\wavdetect}{\textit{wavdetect}}

\begin{document}
%
\title{X-ray Astronomical Point Sources Recognition Using Granular Binary-tree SVM}

\author{
  Zhixian Ma\IEEEauthorrefmark{1},
  Weitian Li\IEEEauthorrefmark{2},
  Lei Wang\IEEEauthorrefmark{1},
  Haiguang Xu\IEEEauthorrefmark{2},
  Jie Zhu\IEEEauthorrefmark{1}\\
  \IEEEauthorblockA{
    \IEEEauthorrefmark{1}Department of Electronic Engineering, Shanghai Jiao Tong University, Shanghai, China\\
    \IEEEauthorrefmark{2}Department of Physics and Astronomy, Shanghai Jiao Tong University, Shanghai, China\\
   Email: \{mazhixian, zhujie\}@sjtu.edu.cn
  }
}


\maketitle

\begin{abstract}
 The study on point sources in astronomical images is of special importance, since most energetic celestial objects in the Universe exhibit a point-like appearance.
An approach to recognize the point sources (PS) in the X-ray astronomical images
using our newly designed granular binary-tree support vector machine (GBT-SVM) classifier is proposed.
First, all potential point sources are located by peak detection on the image.
The image and spectral features of these potential point sources are then extracted. Finally, a classifier to recognize the true point sources is build through the extracted features.
Experiments and applications of our approach on real X-ray astronomical images are demonstrated. Comparisons between our approach and other SVM-based classifiers are also carried out by evaluating the precision and recall rates, which prove that our approach is better and achieves a higher accuracy of around $89\%$.
%
\end{abstract}



%
\IEEEpeerreviewmaketitle

\section{Introduction}
\label{sec.intro}
There are various astronomical objects that are radiating at different electromagnetic bands, from the long-wavelength radio band to the high-energy X-ray and $\Gamma$-ray observations.
Among them, the X-ray observation is an important method to study the Universe, and have already revealed to us many exciting discoveries, such as the active galactic nucleus (AGN) and galaxy clusters filled with hot plasma.
And many sources of our interest exhibit a point-like appearance. However, they are very far away and there are diffuse background radiation all over the sky.
In addition, due to the imperfections of the instruments, the observed images are distorted, which can be described as the convolution with instrumental point spread function (PSF) \cite{Freeman2002}.  It should be noted that the PSF varies across the instrument regions.
Therefore, it is a challenge but of great importance to accurately detect these point sources (PS) from the faint observed images.

Since the astronomical sources are very distant and the X-ray radiation flux is very low, the instrument (e.g., CCD) is thus able to measure the position and energy of \emph{every} incoming X-ray photon (i.e., an event).
All the measured events are stored as an event table, from which the spacial image and/or the spectrum of a specific region can be extracted.
Due to the very low radiation flux, the observed image suffers from significant Poisson noises \cite{Freeman2002, Guglielmetti2009}, which makes the PS detection very difficult and error-prone.

Recently, Masias \etal\ reviewed many source (both point-like and extended) detection methods \cite{Masias2012}.  And these methods can be roughly divided into three categories according their underlying techniques:
(1) wavelet \cite{Freeman2002,Broos2010}; (2) matched filters \cite{Vikhlinin1995}; and (3) Bayesian techniques \cite{Guglielmetti2009}.
Although every method has its own strengthens and applicabilities,
however, most of them only exploit the spacial information of the image, and just ignore the additional spectral information.
We argue that the available spectral information should be utilized together with the spacial information, in order to further improve the source detection ability.

In this paper, we propose a new classifier based on the support vector machine (SVM) to recognize the potential PS, which takes advantage of available information from both the spacial and the spectral domain.
All potential PS are first located by performing the peak detection on the spacial image.  Then, both the spacial and spectral features are extracted and used to recognize the true PS by our SVM-based classifier.

The SVM has been proved as an excellent algorithm to solve the classification problems \cite{Vapnik1995} and has been widely used in various areas \cite{Tang2004,Tang2006,Fei2006}.  Many algorithms based on the SVM are also proposed, such as Granular SVM (GSVM) \cite{Tang2004,Tang2006}, and Binary-Tree SVM (BTS)\cite{Fei2006}.  Among them, the BTS is used for multi-class problems, while the GSVM can handle the imbalanced training sets well.

With regard to the X-ray astronomical images, they are usually very sparse, i.e., the samples of backgrounds are far more than the number of PS.  Besides, the PS should be further divided into bright and faint classes, because the faint ones are almost as faint as the backgrounds.
Consequently, a classifier to solve both the imbalanced training sets and multi-class classification problem is required.
Thus we design our new granular binary-tree SVM (GBT-SVM) classifier by integrating both the GSVM and BTS.

The rest of the paper is organized as follows. In Sec.~\ref{sec.ps.bkg.prop}, we describe the properties and spectral features of the point sources and backgrounds. In Sec.~\ref{sec.Model}, our proposed GBT-SVM classifier is explained in detail. After that, we briefly describe the peak detection approach used to locate the potential PS in Sec.~\ref{sec.PotPS}. Then in Sec.~\ref{sec.Exp}, experiments are carried out using the real X-ray astronomical images. Finally, we conclude in Sec.~\ref{sec.Conclusion} with some discussions and outlooks.

\section{Point Source and Background Properties}
\label{sec.ps.bkg.prop}
The PS have different spacial and spectral properties compared to the backgrounds, thus they can be located in the image, and can be further recognized from backgrounds.
In this section, we introduce both the spacial and spectral properties of the PS and backgrounds, and then define the features used for recognition.

\subsection{Spacial domain}
\label{sec.spacial.prop}
A typical PS appears as an compact elliptical blob on the image that is brighter than its surrounding background.
Its elliptical shape is due to the convolution with the instrumental PSF, which is generally elliptical \cite{Freeman2002}.
And a PS usually have only one major peak at the centroid.

As for the background, it is generally uniformly distributed and obeys the Poisson distribution.  There are many minor peaks within the background region without any major peak.
However, there may exists extended source in the image which is much brighter than the plain background.  Since it is also uniform on small scales of our interest, and it is more similar to the background compared to the PS, we regard the extended source as the \emph{bright background}.

\begin{figure}[t]
\centering
\includegraphics[width=0.48\textwidth]{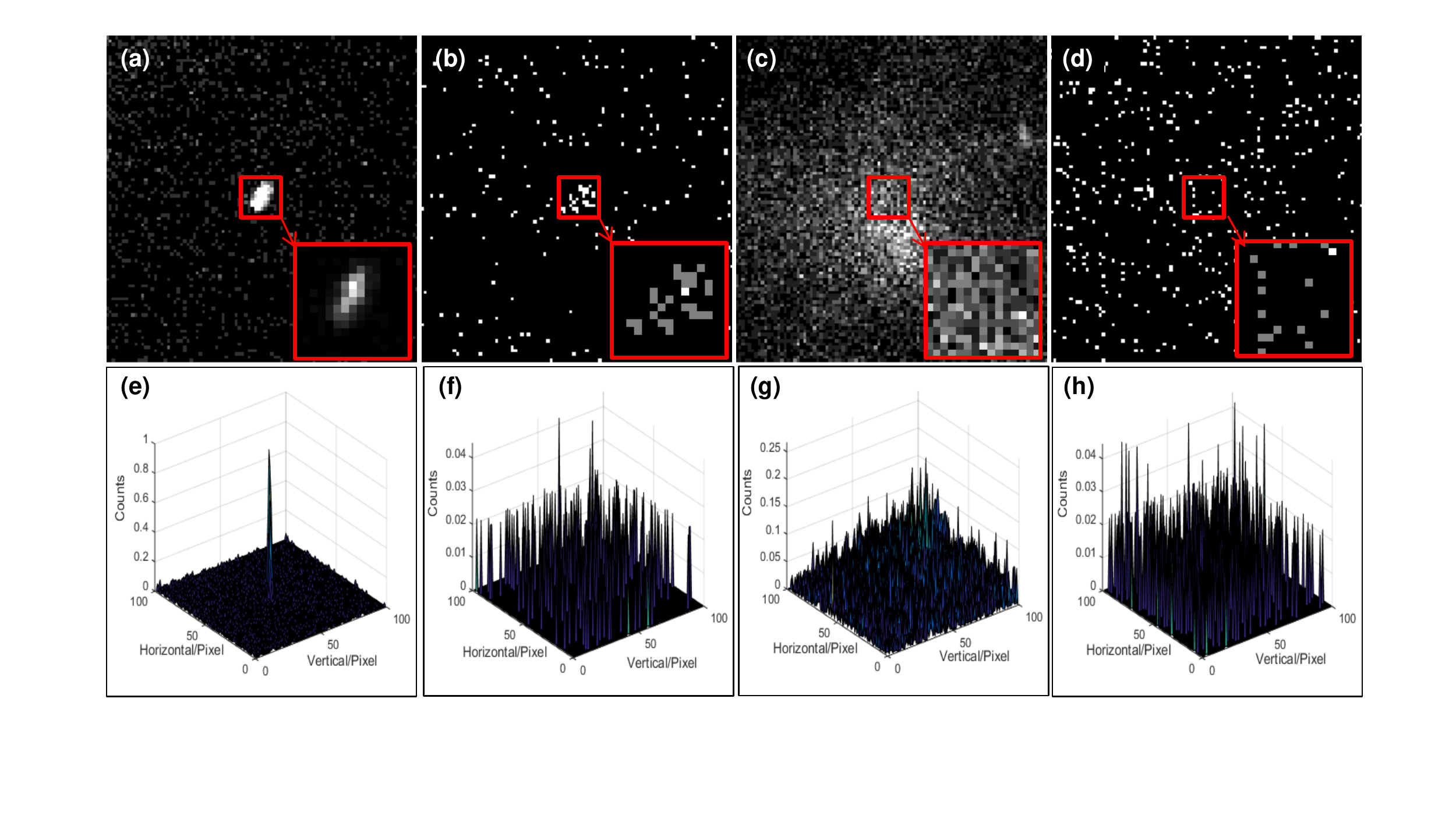}
  \caption{Example images cropped from the Chandra observation of HCG 62 (observation ID: 921) showing the following 4 classes of objects: (a) bright PS, (b) faint PS, (c) bright background (i.e., extended source) and (d) faint background. And the bottom row  shows the corresponding three-dimensional views.}
\label{fig.Exp}
\end{figure}

Therefore, we have following 4 class of objects in our work:
(1) bright PS; (2) faint PS; (3) bright background (i.e., extended source) and (4) faint background.
Fig.~\ref{fig.Exp} shows an example of these 4 class of objects.

\subsection{Spectral domain}
\label{sec.spectral.prop}
\begin{figure}[t]
  \centering
  \subfigure[]{
    \label{fig.SpecCmpAll}
    \includegraphics[width=0.45\textwidth]{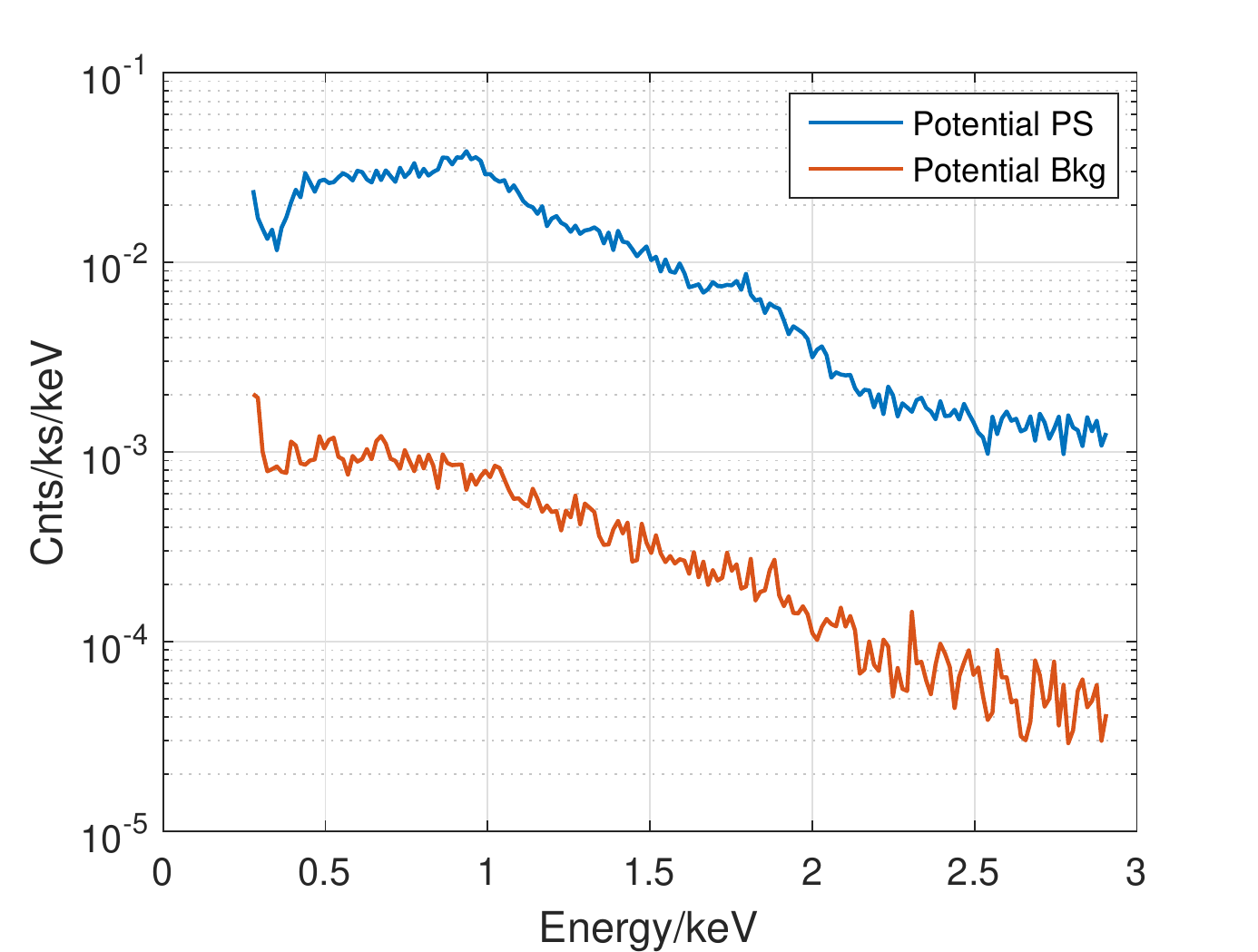}}
  \subfigure[]{
    \label{fig.SpecCmpPotPS}
    \includegraphics[width=0.45\textwidth]{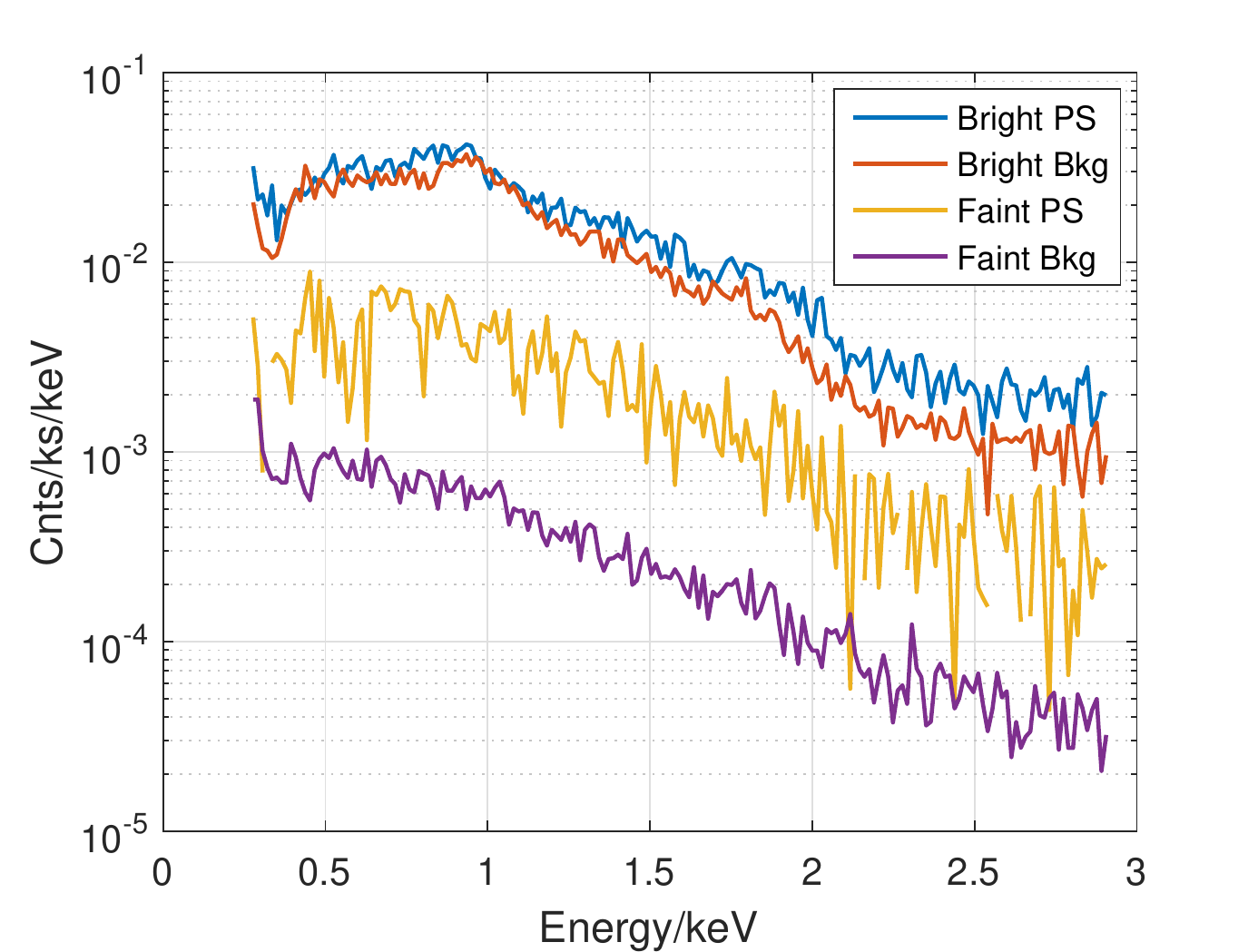}}
  \caption{
    Different spectral intensities and shapes among different class of PS and background.
    \textbf{(a)} Spectra of the potential PS and background (corresponding to the children in the 1$^{\mathrm{st}}$ level of the binary tree as shown in Fig.~\ref{fig.BTSVM});
    \textbf{(a)} Spectra of the 4 different classes of objects (corresponding to the children in the 2$^{\mathrm{nd}}$ level of the binary tree).}
  \label{fig.SpecFeature}
\end{figure}

A spectral describes the energy distribution of the X-ray photons that extracted from the region of interest, which can further reflect the internal physical processes of the origin source.
As for the \emph{Chandra}\footnote{\emph{Chandra} X-ray Observatory: \url{http://cxc.harvard.edu/}} X-ray observations, we extract the spectrum within energy range of 0.5 keV -- 3.0 keV for this work.
Generally, the PS is brighter than the background, thus it also has higher spectral intensities.
In addition, the PS and background have different spectral shape due to there different radiation origins, as shown in Fig.~\ref{fig.SpecCmpAll}.
Therefore, the spectral properties can be utilized to recognize the PS from background.

\begin{figure}[htb]
\centering
\includegraphics[width=0.45\textwidth]{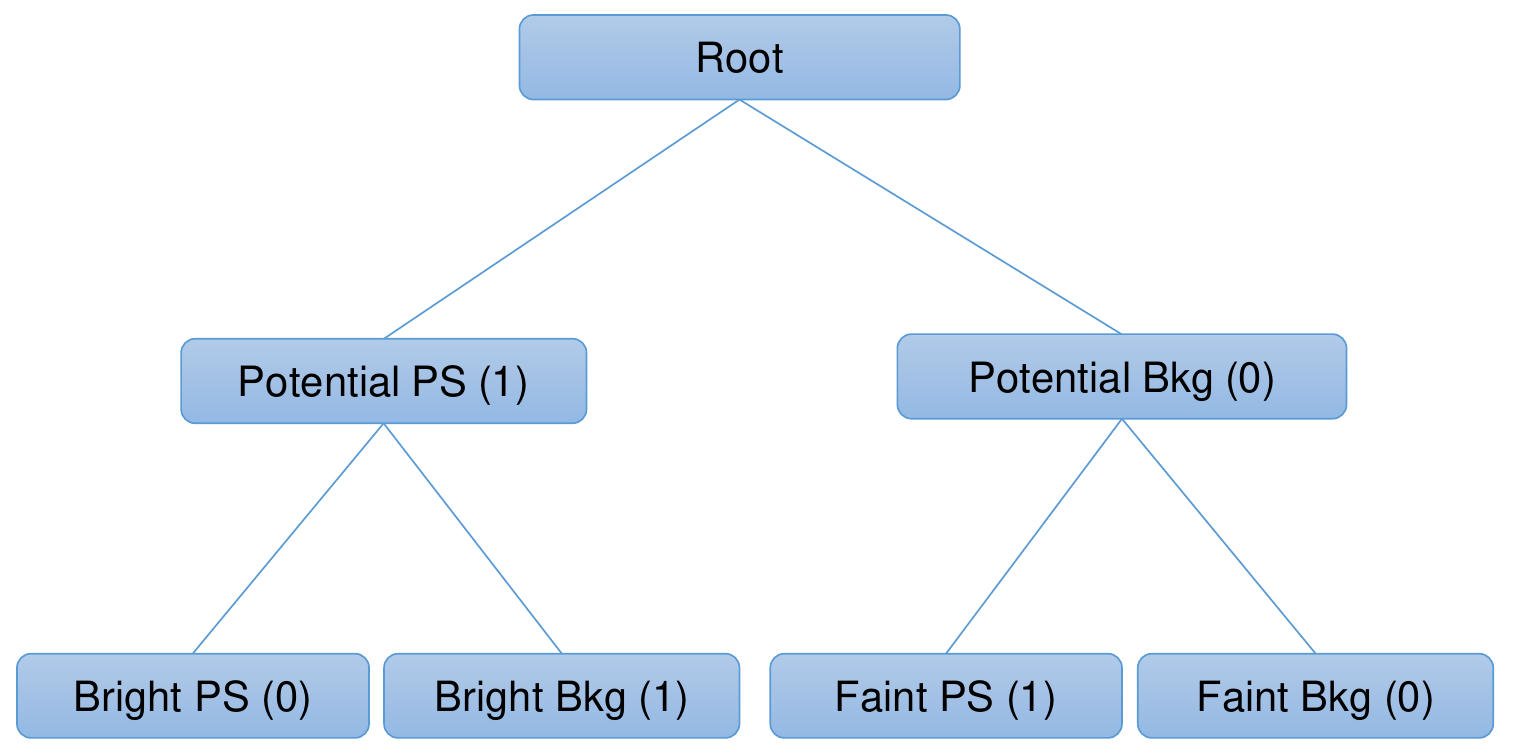}
\caption{A frame of the binary-tree structure SVM classifier.}
\label{fig.BTSVM}
\end{figure}

\subsection{Feature vector}
\label{sec.FtVec}
In order to classify the 4 different classes of objects, we adopt a two-level binary tree for our GBT-SVM, as illustrated in Fig.~\ref{fig.BTSVM}.

The extracted spectrum of an object can be just used as the feature vector for our classifier, since it is already in vector format.
In addition, we define the following 4 spacial features:
(1) counts per pixel ($F_{\mathrm{cpp}}$);
(2) peak-to-average ratio ($F_{\mathrm{par}}$);
(3) variance ($F_{\mathrm{var}}$);
and (4) number of peaks ($F_{\mathrm{nop}}$).
And the features (1)--(3) are defined as:
\begin{align}
  F_{\mathrm{cpp}} & = \frac{1}{mn}\sum^m_{i=1}{\sum^n_{j=1}{R(x_i,y_j)}}, \\
  F_{\mathrm{par}} & = \frac{R(x_p,y_p)}{F_{\mathrm{cpp}}},  \\
  F_{\mathrm{var}} & = \frac{1}{mn}\sum^m_{i=1}{\sum^n_{j=1}{[R(x_i,y_j) - F_{\mathrm{cpp}}]^2}},
\end{align}
where $R$ represents the region of the object of interest, $(x_p,y_p)$ is the coordinates of the peak, and $m,n$ are the rows and columns of matrix $R$.

Then, we combine the above 4 spacial features with the spectral features to derive the final feature vector to be used in our GBT-SVM classifier:
\begin{equation}
  F = ( F_{\mathrm{spec}}, F_{\mathrm{cpp}}, F_{\mathrm{par}}, F_{\mathrm{var}}, F_{\mathrm{nop}}),
\end{equation}
where $F_{\mathrm{spec}}$ is the spectral feature of the object.

\section{GBT-SVM model}
\label{sec.Model}
To tackle the multi-class problem described above, we design a new SVM-based classifier, namely GBT-SVM,
which consists of three submodels: (1) classifier for the potential PS and background; (2) classifier for bright PS and background; and (3) classifier for faint PS and background.
In this section, we first briefly introduce the basically SVM model of our classification problem, then describe the strategy to deal with the imbalanced dataset by adopting the thinking of granular computing and sampling. Finally the GBT-SVM model is explained.

\subsection{Basic SVM model}
\label{sec.BasicSVM}
The spacial and spectral features of each PS and background object are extracted as described in Sec. \ref{sec.FtVec},
and are represented as
\begin{equation}
\label{eq.Sample}
  S =\{ (\bm{x_i},y_i) \}, i = 1,\cdots,M,
\end{equation}
where $S$ is the sample feature set, $\bm{x_i},y_i$ are the $i^\mathrm{th}$ sample's feature vector and its classification label, and $M$ is the amount of samples.

Then the classifier is defined as follows,
\begin{equation}
  f(\bm{x}) = \bm{w}^{T}{\phi(\bm x)} + b,
\label{eq.classifier}
\end{equation}
where $\bm{w}$ is the weight vector, and $b$ is the bias, and $\phi(\bm{x})$ is a mapping function that maps the feature vector $\bm{x}$ to a higher dimensional linear space for easier classification.

And the solution of parameters $\bm{w}$ and $b$ can be derived by solving the following optimization problem, where the soft margin strategy with loss function is also considered (see Eq.~\ref{eq.OptEq}).
\begin{align}
\label{eq.OptEq}
  \min_{\bm{w},b} &\quad \frac {1}{2} {\|\bm{w}\|}^2 + C \sum^{M}_{i=1}\xi_i\notag \\
  \mathbf{s.t.} &\quad  y_i(\bm{w}^{t}\phi(\bm{x_i}) + b) \geq 1 - \xi_i,  \notag \\
  &\quad \xi_i  \geq 0, i = 1,2,\cdots,m,
\end{align}
where $C$ is the trade-off between the structural risk (target) and the empirical risk (miss probability)\cite{Chang2011}. $\xi_i, i=1,2,\cdots,M$ are the slack variables, which are defined as:
\begin{equation}
\label{eq.Slack}
  \xi_i = \max (0,1 - y_i(\bm{w}^T\phi(\bm{x_i})+b))
\end{equation}
To solve Eq.\ref{eq.OptEq}, the Lagrange function is defined.
\begin{align}
\label{eq.Lagrange}
  L(\bm{w},b,\bm{\alpha},\bm{\xi},\bm{\mu}) = & \frac {1}{2} {\|\bm{w}\|}^2 + C \sum^{M}_{i=1}\xi_i  \notag \\
    & + \sum^{M}_{i = 1}\alpha_i(1-\xi_i-y_i(\bm{w}^{T}\phi(\bm{x_i})+b))  \notag \\
    & - \sum^{M}_{i=1}\mu_i \xi_i,
\end{align}
where $\alpha_i, \mu_i$ are the Lagrange multipliers. Let the partial derivation of $L$ with respect to parameters $\bm{w},b,\xi_i$ be zero, then the above equation is rearranged to its dual form:
\begin{align}
\label{eq.LagDual}
  \max_{\bm{\alpha}} &\quad \sum^{M}_{i=1}{\alpha_i} - \frac{1}{2} \sum^{M}_{i=1}{\sum^{M}_{j=1}{\alpha_i \alpha_j y_i y_j \phi(\bm{x_i})^T \phi(\bm{x_j})}}, \notag \\
  \mathbf{s.t.}  &\quad  \sum^{M}_{i=1}{\alpha_i}y_i = 0,  \notag \\
  & \quad 0\leq \alpha_i \leq C, i = 1,2,\cdots,M.
\end{align}
Then we solve Eq.~\ref{eq.classifier} by calculating $\alpha_i,i=1,2,\cdots,M$, and the result is:
\begin{equation}
  f(\bm{x}) = \sum^M_{i = 1}{\alpha_i y_i[ \phi(\bm{x_i})^T \phi(\bm{x})]} + b,
\end{equation}
where $\phi(\bm{x_i})^T \phi(\bm{x})$ can be calculated by the kernel function $\kappa(\bm{x_i},\bm{x_j})$. In this work, the RBF (Radical Basis Function) kernel is used \cite{Chang2011}.

\subsection{Imbalanced dataset}
\label{sec.GSVM}
Since the amount of faint background samples is far more than the point and extended sources, which leads to an imbalanced sample set.  Tang \etal{}  proposed a method namely Granular SVM (GSVM), which separates the majority sample set into multiple subsets (i.e., information granules) so as to achieve a better hyperplane \cite{Tang2006,Tang2009}.  In this work, we take advantage of this method to deal with the imbalanced sample sets.

Taking the classification between the potential PS and background as an example, it compares sample amounts among the classes and choose the larger class as the major class.  In this example, the potential background is the major class, while the potential PS is the minor one.  Then the number of granular sets is determined by:
\begin{equation}
\label{eq.GraNum}
     N_{\mathrm{gra}} = \lfloor \frac{N_{\mathrm{maj}}} {N_{\mathrm{min}}} \rfloor,
\end{equation}
where $\lfloor \cdot \rfloor$ means rounding down.

After that, the major sample set is separated into $N_{gra}$ subsets. In this work, the samples are acquired from different observations of different exposure times. Thus, we take advantages of Tang \etal{}'s methods of under-sampling \cite{Tang2006}, and generate the granular sets by uniformly sampling on the major sample sets so as to cover all the observations.
\begin{align}
\label{eq.SubSpl}
  S^{k}_{\mathrm{maj}} &= \{S^{k+iN_{\mathrm{gra}}}_{\mathrm{maj}}\}, \notag \\
  i & = 1,2,\cdots,N_{min}, k=1,2,\cdots,N_{gra}.
\end{align}

Finally, there generates $N_\mathrm{gra}$ sub training sets. As for each subset, it combines a granular  of major samples and the whole minor samples, and respective SVM model can be trained and obtained. 
With those submodels, the class label of a sample can be decided by a voting strategy, i.e., the label with most tickets wins.

\subsection{GBT-SVM}
\label{set.GBTSVM}
We build our GBT-SVM classifier by combining the basic SVM and the granular sets.
Our classifier has a binary tree structure, and the whole model is divided into three submodels.
In each submodel, the major sample is evaluated and separated into granular subsets, and the number of subsets is calculated by Eq.~\ref{eq.GraNum}. Then the classifier is obtained as follows,
\begin{equation}
  C = C_1 \oplus (C_{\mathrm{2L}} \cup C_{\mathrm{2R}}),
\end{equation}
where $C_1$ is the GSVM classifier of 1$^\mathrm{st}$ level (i.e., potential PS and background), $C_{\mathrm{2L}}$ and $C_{\mathrm{2R}}$ are the GSVM classifiers of leaves in our binary-tree model. And total number of submodels in the classifier is:
\begin{equation}
\label{Eq.ModelNum}
  N_{C} = N^1_{\mathrm{gra}} + N^{\mathrm{2L}}_{\mathrm{gra}} + N^{\mathrm{2R}}_{\mathrm{gra}}.
\end{equation}

In addition, the 4 classes of objects are encoded based on their labels so as to estimate whether a sample is PS or background.
We use three bits to encode the information of each object.
The first two bits are labels of the two levels in our binary tree as Fig.~\ref{fig.BTSVM} shows.
And the third bit is the decision label which is the \textit{XOR} of the first two bits (See Tab.~\ref{tab.code}).

The GBT-SVM classification algorithm is displayed in Alg.~\ref{alg.GBTAlg}.

\begin{table}[h]
\renewcommand{\arraystretch}{1.3}
\caption{Code table for the leaves in the binary-tree structure.}
\label{tab.code}
\centering
\small
\begin{tabular}{|c|c|c|c|}
\hline
Region type & $\mathrm{Label_1}$ & $\mathrm{Label_2}$ & Decision~label \\ \hline
Bright PS & 1 & 0 & 1 \\ \hline
Bright Bkg & 1 & 1 & 0 \\ \hline
Faint PS & 0 & 1 & 1 \\ \hline
Faint Bkg & 0 & 0 & 0 \\ \hline
\end{tabular}
\end{table}

\begin{algorithm}[t]
  \caption{GBT-SVM decision algorithm}
  \label{alg.GBTAlg}
  \small
  \begin{algorithmic}[1]
    \STATE \textbf{Input:} $Sample$ and \textbf{Load:} $Classifiers$
    \STATE $Predict_{1} = zeros(1, N^{1}_{\mathrm{gra}})]$
    \FOR {$i = 1 : N^{1}_{\mathrm{gra}}$}
      \STATE $Predict_{1}(i) = predict(Sample,Classifiers.L_{1}(i))$
    \ENDFOR
    \STATE $Label_{1} = vote(Predict_{1})$
    \IF {$Label_1 == 1$}
      \STATE $Predict_{\mathrm{2L}} = zeros(1, N^{\mathrm{2L}}_{\mathrm{gra}})]$
      \FOR {$j = 1 : N^{\mathrm{2L}}_{\mathrm{gra}}$}
        \STATE $Predict_{\mathrm{2L}}(j) = predict(Sample,Classifiers.L_{\mathrm{2L}}(j))$
      \ENDFOR
      \STATE $Label_{2} = vote(Predict_{\mathrm{2L}})$
    \ELSE
      \STATE $Predict_{\mathrm{2R}} = zeros(1, N^{\mathrm{2R}}_{\mathrm{gra}})]$
      \FOR {$j = 1 : N^{\mathrm{2R}}_{\mathrm{gra}}$}
        \STATE $Predict_{\mathrm{2R}}(j) = predict(Sample,Classifiers.L_{\mathrm{2R}}(j))$
      \ENDFOR
      \STATE $Label_{2} = vote(Predict_{\mathrm{2R}})$
    \ENDIF
    \STATE $FinalLabel = xor(Label_{1},Label_{2})$
    \IF {$FinalLabel == 1$}
      \STATE \textbf{Output:} The sample is a point source.
    \ELSE
      \STATE \textbf{Output:} The sample is a background.
    \ENDIF
  \end{algorithmic}
\end{algorithm}

\section{Potential PS localization}
\label{sec.PotPS}
In this section, the approach employed to locate all the potential PS in described.
A PS generally has a major peak on the image, as mentioned in Sec~\ref{sec.spacial.prop}, so it is intuitive to adopt the peak detection method.

The background noise is suggested to subject to the Poisson distribution \cite{Guglielmetti2009}. And for the parameter $\lambda$ in such distribution, it can be estimated without bias using the average of the group of samples \cite{Tinmmermann1999}:
\begin{equation}
  \hat{\lambda} = \bar{I} = \frac {\sum_{i=1}^M \sum_{j=1}^N I_{i,j}} {MN},
\label{eq.EstPos}
\end{equation}
where $\hat{\lambda}$ is the estimated value, and $I$ is the image matrix.

Thus, the raw image can be preprocessed to reduce the noise.  The parameter $\lambda$, is estimated and set as a threshold, pixels with counts less than which are set to zero (See Eq.~\ref{eq.BkgThrs}).
\begin{equation}
\label{eq.BkgThrs}
I_r(i,j) = \begin{cases}
0, & I(i,j) \leq \hat\lambda; \\
I(i,j), & I(i,j) > \hat\lambda, \\
\end{cases}
\end{equation}
where $\{ (i,j);i=1,\cdots,M; j=1, \cdots,N \}$ are the pixel coordinates in the image $I$, and $I_r$ is the noise-reduced image.

Finally, peaks in the preprocessed image are located and listed as potential point sources.
In our work, to reduce the complexities, we do not detect peaks by covering all pixels.
Instead, pixels of the two dimensional matrix are sorted in descending order, and only pixels with values greater than a threshold are extracted as PS centers.

However, as for the extended sources (i.e., bright backgrounds), they often have spread  bright pixels. Thus local maxima in these regions are not as significant as point sources.
To solve this problem, the neighbors around a maximum pixel are considered. If there are pixels whose values equal or approach to the maximum, it will be eliminated from the PS list.

\section{Experiments and Result}
\label{sec.Exp}
Experiments on real X-ray astronomical observations were carried out to demonstrate the performance of our proposed GBT-SVM classifier. And comparisons of our approach with GSVM, and BTSVM were also performed.

\subsection{Datasets}
\label{sec.Data}
All the datasets in our work were obtained from the \textit{Chandra Data Archive}\footnote{Chandra Data Archive: \url{http://cxc.harvard.edu/cda/}} and processed using the CIAO\footnote{CIAO: \url{http://cxc.harvard.edu/ciao/}} software v4.4 by following the official guide \cite{Fruscione2006}.
After manually filtering, 25 observations were selected, among which 20 observations were chosen as the training datasets while the remaining 5 were used as the testing datasets.

Each observation averagely has 20 point sources and the average exposure time is about 41.62 ks.  To evaluate the performance of our approach, all point sources in the raw images were detected with \wavdetect{}\cite{Freeman2002} provided by the CIAO software and then visually checked. They were then set as the \textit{reference group}.

For each training observation, we randomly selected 150 faint backgrounds, 30 bright backgrounds, and the bright and faint PS were manually distinguished according to their brightness and spectrum in the reference group. Then the feature vectors of each sample were generated as explained in Sec.~\ref{sec.FtVec}.

As for the testing dataset, the potential point sources were detected with peak detection method as described in Sec.~\ref{sec.PotPS}, and then the corresponding features were extracted.

\subsection{Evaluation strategy}
The accuracy $R_A$ of our PS recognition approach is defined by the combination of true positive (TP) and false negative (FN) measurements \cite{Masias2012, Fawcett2006}:
\begin{align}
\label{eq.Accuracy}
  R_A = \frac{N_{\mathrm{TP}} + N_{\mathrm{FN}}} {N_S},
\end{align}
where $N_S$ is the number of samples, $N_\mathrm{TP}$ represents the true PS our approach recognized, and $N_\mathrm{FN}$ is the number of true backgrounds discarded.

In addition, to evaluate the generalization abilities of our GBT-SVM classifier, and compare it with other SVM based methods, two famous measurements are utilized, i.e., the \textit{precision} and \textit{recall} rates \cite{Harrington2012}.
As for a good or generalized classifier, both of the precision and recall rates should be large enough. The two measurements are defined as:
\begin{align}
  P & = \frac {\overline{TP}} {\overline{TP} +\overline{FP}}, \\
  R & = \frac {\overline{TP}} {\overline{TP} + \overline{FN}},
\end{align}
where $P$ and $R$ are precision and recall rates, respectively; $\overline{TP}$, $\overline{FP}$, $\overline{TN}$, and $ \overline{FN}$ are the true positive, false positive, true negative and false negative detections averaged over all the tree-structured submodels, respectively.

\subsection{Experiments and comparisons}
\label{sec.ExpAndCmp}
\begin{table}[t]
\renewcommand{\arraystretch}{1.4}
  \caption{Results of the potential PS located by the peak detection approach. $N_{\mathrm{ref}}$ is number of true PS according to the reference group. $N_T$ and $N_F$ are the number of true PS and false PS detected by our approach and false detected PS, respectively.}
\label{tab.PotPS}
\centering
\footnotesize
\begin{tabular}{|c|c|c|c|c|}
\hline
Name (ObsID) & $N_\mathrm{ref}$ & $N_T$ &  $N_F$ & Accuracy ($\%$)\\
\hline
3C 186 (9774) & 42 &40&13& 75.47 \\
\hline
MACS J2140.2-2339 (4974) &15&15 & 4 & 78.95 \\
\hline
NGC 6482 (3218) & 13 & 13 & 7&65.00 \\
\hline
NGC 7619 (3955) &  21 & 20 & 4 & 83.33 \\
\hline
RCS J1107.3-0523 (5825) & 25 & 25 &  2 & 92.59 \\
\hline
\end{tabular}
\end{table}

\begin{table}[t]
\renewcommand{\arraystretch}{1.4}
\caption{Comparisons among the classification results of GSVM, BT-SVM, and GBT-SVM. }
\label{tab.CmpRst}.
\centering
\footnotesize
\begin{tabular}{|c|c|c|c|c|}
\hline
Name (ObsID)&Approaches& $P~(\%)$ & $R~(\%)$ & Accuracy$(\%)$  \\
\Xhline{1pt}
\multirowcell{3}{3C 186 (9774)}&GSVM & 61.90 & 72.22 &33.96  \\
\cline{2-5} & BT-SVM & 73.81 & 93.94 & 62.26 \\
\cline{2-5} & \textbf{GBT-SVM}  & 81.40 & 87.50 & \textbf{75.47} \\
\Xhline{1pt}
\multirowcell{3}{MACS J2140.2-\\2339 (4974)}&GSVM  & 81.25 & 92.86 & 73.68 \\
\cline{2-5} & BT-SVM & 81.25 & 92.86 & 73.68 \\
\cline{2-5} & \textbf{GBT-SVM}  & 82.35 & 93.33 & \textbf{78.95} \\
\Xhline{1pt}
\multirowcell{3}{NGC 6482 (3218)}&GSVM  & 65.00 & 100.00 & 65.00\\
\cline{2-5} & BT-SVM & 68.42 & 92.86 & 70.00 \\
\cline{2-5} & \textbf{GBT-SVM}  & 81.25 & 76.47 & \textbf{85.00} \\
\Xhline{1pt}
\multirowcell{3}{NGC 7619 (3955)}&GSVM  & 83.33 & 100.00 & 83.33 \\
\cline{2-5} & BT-SVM & 83.33 & 100.00 & 83.33  \\
\cline{2-5} & \textbf{GBT-SVM}  & 90.91 & 90.91 & \textbf{91.67} \\
\Xhline{1pt}
\multirowcell{3}{RCS J1107.3-0523 \\(5825)}&GSVM  & 95.24 & 95.24 & 77.78 \\
\cline{2-5} & BT-SVM & 92.31 & 100.00 & 88.89 \\
\cline{2-5} & \textbf{GBT-SVM}  & 96.00 & 100.00 & \textbf{92.59} \\
\hline
\end{tabular}
\end{table}

Above all, the peak detection approach described in Sec.~\ref{sec.PotPS} were applied to generate potential PS lists in the X-ray images. And the results were listed in Tab~\ref{tab.PotPS}).
Compared with the reference PS,  our peak detection algorithm located nearly all of the true point sources, but with some spurious detections.

Therefore, we apply our GBT-SVM classifier to recognize the true PS and to discard  the spurious PS for each test observations.  The classifier was already trained with the training datasets.

The results are displayed in Tab.~\ref{tab.CmpRst}. And the GSVM and BT-SVM classifiers were also applied to recognize the true PS for comparisons.
It can be found that, our proposed GBT-SVM classifier achieved the highest detection accuracy and thus had the best performance among all of the methods.
Besides, the precision and recall measurements of the GBT-SVM were also very promising.

In addition, as for GSVM and  BTSVM,  the accuracy rates of some observations  are even less than the accuracy rates only after peak detection.  In our opinion, it is because some true point sources were wrongly classified as bright backgrounds, as well as the spurious PS were misjudged as point sources.  

Finally, we also combined all the 5 test observations and carried out the comparison, and the detection accuracy of GSVM, BT-SVM and GBT-SVM were $75.56\%$, $80\%$ and $88.89\%$, respectively.

\section{Conclusion}
\label{sec.Conclusion}
In this paper we propose a new point source recognition approach for the X-ray astronomical observations. The potential PS are first located by the peak detection approach, and then a newly designed Granular Binary-tree SVM (GBT-SVM) classifier is trained to recognize the true PS with the spurious PS discarded.
Our approach not only utilize the spacial features, but also fully exploit the available spectral features of the PS and background.
And the comparison results presented in Sec.~\ref{sec.Exp} highlight that our approach is accurate and has good generalization abilities.

We also compare our classification approach to the other SVM-based approaches, i.e., GSVM and BT-SVM. It shows that our classifier achieved a significantly higher detection accuracy, which proves that our GBT-SVM classifier is accurate and is very suitable for the X-ray PS recognition.

In the further work, we are planning to find approaches to accurately determine the outlines of the PS, so that we are able to better analyze these objects.

\section*{Acknowledgment}
This work is supported by the National Natural Science Foundation of China (grant Nos. 11433002, 61271349, and 61371147), and Shanghai Academy of Spaceflight Technology (grant No. SAST2015039).



\bibliographystyle{IEEEtran}
\bibliography{Ref}

\begin{thebibliography}{10}
\providecommand{\url}[1]{#1}
\csname url@samestyle\endcsname
\providecommand{\newblock}{\relax}
\providecommand{\bibinfo}[2]{#2}
\providecommand{\BIBentrySTDinterwordspacing}{\spaceskip=0pt\relax}
\providecommand{\BIBentryALTinterwordstretchfactor}{4}
\providecommand{\BIBentryALTinterwordspacing}{\spaceskip=\fontdimen2\font plus
\BIBentryALTinterwordstretchfactor\fontdimen3\font minus
  \fontdimen4\font\relax}
\providecommand{\BIBforeignlanguage}[2]{{%
\expandafter\ifx\csname l@#1\endcsname\relax
\typeout{** WARNING: IEEEtran.bst: No hyphenation pattern has been}%
\typeout{** loaded for the language `#1'. Using the pattern for}%
\typeout{** the default language instead.}%
\else
\language=\csname l@#1\endcsname
\fi
#2}}
\providecommand{\BIBdecl}{\relax}
\BIBdecl

\bibitem{Freeman2002}
P.~E. Freeman, V.~Kashyap, R.~Rosner, and D.~Q. Lamb, ``A wavelet-based
  algorithm for the spatial analysis of poisson data,'' \emph{The Astrophysical
  Journal}, vol. 138, pp. 185--218, 2002.

\bibitem{Guglielmetti2009}
F.~Guglielmetti, R.~Fischer, and V.~Dose, ``Background-source separation in
  astronomical images with bayesian probability theory --- i. the method,''
  \emph{Monthly Notices of the Royal Astronomical Society}, vol. 396, no.~1,
  pp. 165--190, 2009.

\bibitem{Masias2012}
M.~Masias, J.~Freixenet, X.~Llad\'o, and M.~Peracaula, ``A review of source
  detection approaches in astronomical images,'' \emph{Monthly Notices of the
  Royal Astronomical Society}, vol. 422, no.~2, pp. 1674--1689, 2012.

\bibitem{Broos2010}
P.~S. Broos, L.~K. Townsley, E.~D. Feigelson, K.~V. Getman, F.~E. Bauer, and
  G.~P. Garmire, ``Innovations in the analysis of chandra-acis observations,''
  \emph{The Astrophysical Journal}, vol. 714, no.~2, p. 1582, 2010.

\bibitem{Vikhlinin1995}
A.~Vikhlinin, W.~Forman, C.~Jones, and S.~Murray, ``Rosat extended medium-deep
  sensitivity survey: Average source spectra,'' \emph{The Astrophysical
  Journal}, vol. 451, p. 564, 1995.

\bibitem{Vapnik1995}
C.~Cortes and V.~Vapnik, ``Support-vector networks,'' \emph{Machine learning},
  vol.~20, no.~3, pp. 273--297, 1995.

\bibitem{Tang2004}
Y.~Tang, B.~Jin, Y.~Sun, and Y.-Q. Zhang, ``Granular support vector machines
  for medical binary classification problems,'' in \emph{Computational
  Intelligence in Bioinformatics and Computational Biology, 2004. CIBCB'04.
  Proceedings of the 2004 IEEE Symposium on}.\hskip 1em plus 0.5em minus
  0.4em\relax IEEE, 2004, pp. 73--78.

\bibitem{Tang2006}
Y.~Tang and Y.-Q. Zhang, ``Granular svm with repetitive undersampling for
  highly imbalanced protein homology prediction,'' in \emph{Granular Computing,
  2006 IEEE International Conference on}.\hskip 1em plus 0.5em minus
  0.4em\relax IEEE, 2006, pp. 457--460.

\bibitem{Fei2006}
B.~Fei and J.~Liu, ``Binary tree of svm: a new fast multiclass training and
  classification algorithm,'' \emph{Neural Networks, IEEE Transactions on},
  vol.~17, no.~3, pp. 696--704, 2006.

\bibitem{Chang2011}
C.-C. Chang and C.-J. Lin, ``Libsvm: a library for support vector machines,''
  \emph{ACM Transactions on Intelligent Systems and Technology (TIST)}, vol.~2,
  no.~3, p.~27, 2011.

\bibitem{Tang2009}
Y.~Tang, Y.-Q. Zhang, N.~V. Chawla, and S.~Krasser, ``Svms modeling for highly
  imbalanced classification,'' \emph{Systems, Man, and Cybernetics, Part B:
  Cybernetics, IEEE Transactions on}, vol.~39, no.~1, pp. 281--288, 2009.

\bibitem{Tinmmermann1999}
K.~Tinmmermann and R.~Nowak, ``Multiscale modeling and estimation of poisson
  processes with application to photon-limited imaging,'' \emph{IEEE
  Transactions on Information Theory}, vol.~45, no.~3, pp. 846--862, April
  1999.

\bibitem{Fruscione2006}
A.~Fruscione, J.~C. McDowell, G.~E. Allen, N.~S. Brickhouse, D.~J. Burke, J.~E.
  Davis, N.~Durham, M.~Elvis, E.~C. Galle, D.~E. Harris \emph{et~al.}, ``Ciao:
  Chandra's data analysis system,'' in \emph{SPIE Astronomical Telescopes+
  Instrumentation}.\hskip 1em plus 0.5em minus 0.4em\relax International
  Society for Optics and Photonics, 2006, pp. 62\,701V--62\,701V.

\bibitem{Fawcett2006}
T.~Fawcett, ``An introduction to roc analysis,'' \emph{Pattern recognition
  letters}, vol.~27, no.~8, pp. 861--874, 2006.

\bibitem{Harrington2012}
P.~Harrington, \emph{Machine learning in action}.\hskip 1em plus 0.5em minus
  0.4em\relax Manning, 2012.

\end{thebibliography}

\end{document}